\documentclass[3p,times]{elsarticle}
\usepackage{framed,multirow}
\usepackage{amssymb}
\usepackage{latexsym}
\usepackage{url}
\usepackage{float}
\usepackage{subfigure}
\usepackage{hyperref}
\usepackage{amsmath}
\usepackage[font=footnotesize,labelfont=bf]{caption}
\usepackage[ruled,vlined,linesnumbered]{algorithm2e}
\usepackage{makecell} 
\usepackage{algpseudocode}
\begin{document}

\begin{frontmatter}
\title{Air Quality Prediction Using LOESS–ARIMA and Multi-Scale CNN–BiLSTM with Residual-Gated Attention}

% \author[a]{Soham Pahari\corref{cor1}}
% \ead{soham.109424@stu.upes.ac.in & paharisoham@gmail.com}

% \author[a]{Sandeep Chand Kumain}
% \ead{sandeep.kumain@ddn.upes.ac.in &  sandeep.chandphdrw@gmail.com}

% \cortext[cor1]{Corresponding author}

% \address[a]{School of Computer Science, UPES, India}

% \fntext[a]{Student, School of Computer Science, UPES, India}

% \fntext[b]{Assistant Professor, School of Computer Science, UPES}

\author[a]{Soham Pahari\corref{cor1}}
\ead{soham.109424@stu.upes.ac.in, paharisoham@gmail.com}

\author[a]{Sandeep Chand Kumain}
\ead{sandeep.kumain@ddn.upes.ac.in, sandeep.chandphdrw@gmail.com}

\cortext[cor1]{Corresponding author}
\address[a]{School of Computer Science, University of Petroleum and Energy Studies (UPES), India}

\begin{abstract}
Air pollution remains a critical environmental and public health concern in Indian megacities such as Delhi, Kolkata, and Mumbai, where sudden spikes in pollutant levels challenge timely intervention. Accurate Air Quality Index (AQI) forecasting is difficult due to the coexistence of linear trends, seasonal variations, and volatile nonlinear patterns. This paper proposes a hybrid forecasting framework that integrates LOESS decomposition, ARIMA modeling, and a multi-scale CNN--BiLSTM network with a residual-gated attention mechanism. The LOESS step separates the AQI series into trend, seasonal, and residual components, with ARIMA modeling the smooth components and the proposed deep learning module capturing multi-scale volatility in the residuals. Model hyperparameters are tuned via the Unified Adaptive Multi-Stage Metaheuristic Optimizer (UAMMO), combining multiple optimization strategies for efficient convergence. Experiments on 2021--2023 AQI datasets from the Central Pollution Control Board show that the proposed method consistently outperforms statistical, deep learning, and hybrid baselines across PM$_{2.5}$, O$_3$, CO, and NO\textsubscript{x} in three major cities, achieving up to 5--8\% lower MSE and higher $R^2$ scores ($>0.94$) for all pollutants. These results demonstrate the framework’s robustness, sensitivity to sudden pollution events, and applicability to urban air quality management.

\end{abstract}


\end{frontmatter}

%\linenumbers

\section{Introduction}

Air pollution has emerged as one of the most pressing environmental and public health challenges in developing nations, with India being no exception\cite{int1}. Major metropolitan cities such as Delhi, Kolkata, and Mumbai frequently rank among the most polluted cities in the world \cite{int3}. Rapid industrialization, explosive population growth, and accelerated urbanization, combined with unregulated vehicular emissions, construction dust, and industrial discharge, have aggravated air quality to alarming levels. \cite{int2}

The Air Quality Index (AQI) provides a standardized measure for evaluating air pollution severity, with four primary pollutants playing a dominant role: particulate matter (PM$_{2.5}$), nitrogen oxides (NOx), carbon monoxide (CO), and ground-level ozone (O$_3$). Prolonged exposure to these pollutants has been strongly associated with severe respiratory illnesses, including asthma, bronchitis, and emphysema, as well as cardiovascular, neurological, and other chronic health complications\cite{int4}.

Accurately forecasting air quality is a highly complex task due to the interplay of numerous linear and nonlinear factors, seasonal variations, and sudden anomalies in pollutant levels. Traditional statistical models, such as ARIMA, capture linear temporal dependencies effectively but struggle with highly nonlinear residual patterns. Conversely, deep learning models like LSTM\cite{ref50} and CNN have demonstrated strong capabilities in modeling complex temporal dependencies but often fail to leverage the structured nature of decomposable time-series components. Many existing hybrid approaches\cite{ref60} fall short because they either:
\begin{enumerate}
    \item Treat the residual component as a black box without explicitly modeling its volatility and multi-scale characteristics, or
    \item Lack a robust mechanism for optimal hyperparameter selection, leading to suboptimal generalization performance.
\end{enumerate}

To address these shortcomings, this paper introduces a multi-stage hybrid forecasting framework that integrates LOESS-based time series decomposition, ARIMA modeling for smooth components, and a multi-scale CNN–BiLSTM network with a residual-gated attention mechanism for volatile residuals. Furthermore, the framework incorporates UAMMO, a unified adaptive metaheuristic optimization algorithm, to fine-tune hyperparameters for optimal predictive performance.

The key contributions of this work are as follows:
\begin{itemize}
    \item Component-wise modeling: Leveraging LOESS decomposition to assign ARIMA for trend and seasonal components, and a deep residual network for noisy components.
    \item Residual-gated attention mechanism: Introducing volatility-aware attention to better capture sudden pollutant spikes and irregularities.
    \item Multi-scale feature extraction: Using multiple convolutional branches to capture both short-term bursts and long-term deviations in residual patterns.
    \item Metaheuristic-driven optimization: Employing UAMMO to efficiently search for optimal hyperparameters across CNN, BiLSTM, and attention layers.
\end{itemize}

The remainder of this paper is organized as follows: Section \ref{sec:rw} reviews related works, Section \ref{sec:pm} details the proposed methodology, Section \ref{sec:exp} presents experimental results, and Section \ref{sec:end} concludes with insights and potential directions for future research.

\section{Related Work}
\label{sec:rw}

Approaches for predicting the Air Quality Index (AQI) can generally be grouped into four categories: (i) statistical models, (ii) machine learning (ML) methods, (iii) deep learning (DL) methods, and (iv) hybrid approaches. Statistical models depend on assumptions regarding the underlying data distribution, aiming to identify causal relationships and estimate unknown parameters. In AQI prediction, these methods often include autoregressive (AR), autoregressive integrated moving average (ARIMA), gray models, and multiple linear regression (MLR). For example, Carbajal-Hernández et al. \cite{ref8} designed an algorithm that evaluates pollution levels through a fuzzy reasoning system to develop a novel air quality index. Zhang et al. \cite{ref9} compared generalized additive models (GAMs) against conventional linear regression techniques. Zhao et al. \cite{ref10} applied an ARIMA model to PM2.5 annual data, confirming via the augmented Dickey–Fuller test that first-order differencing was necessary. A seasonally nonlinear gray model was also proposed to better account for periodic and nonlinear patterns in fluctuating pollution data \cite{ref11}.

Machine learning methods have improved predictive accuracy by leveraging large datasets and overcoming convergence issues. Mehmood et al. \cite{ref12} discussed the transition from conventional approaches to ML-based frameworks. Through trend analysis, ML techniques can also highlight promising research avenues. Usha et al. \cite{ref13} used Neural Networks and Support Vector Machines (SVMs) to improve prediction accuracy and suggested their adaptability for other smart city environments. In another example, Elsheikh \cite{ref14} explored ML applications in friction stir welding for predicting joint quality and failure diagnostics. Ke et al. \cite{ref15} developed an ML-based system that forecasts daily pollutant concentrations using meteorological data, emission levels, and reanalysis datasets, employing an ensemble of five ML models with automated model and hyperparameter selection. Zhang et al. \cite{ref16} investigated ML techniques to predict indoor air quality variations caused by unpredictable factors. Gu et al. \cite{ref17} proposed an interpretable hybrid ML model for PM2.5 prediction that excelled in accuracy, particularly for peak values. More recently, Rakholia et al. \cite{ref18} incorporated meteorology, traffic, industrial and residential pollution levels, spatial data, and time series analysis into a unified air quality prediction framework.

With the advancements in DL, these models have become highly effective in AQI prediction due to their ability to process large datasets and capture complex relationships. Janarthanan et al. \cite{ref19} combined Support Vector Regression (SVR) with LSTM networks to accurately predict AQI, enabling policy decisions such as traffic management and tree planting. Zhang et al. \cite{ref20} examined DL models from temporal, spatial, and spatio-temporal perspectives. Saez et al. \cite{ref21} developed a hierarchical Bayesian spatiotemporal model to achieve accurate predictions even with sparse monitoring stations. Jurado et al. \cite{ref22} utilized convolutional neural networks (CNNs) for fast, precise forecasts using real-time wind, traffic, and urban geometry data. Zhou et al. \cite{ref23} combined CNNs with Gated Recurrent Units (GRUs) to capture both spatial features and temporal dependencies. Mao et al. \cite{ref24} extended LSTM with a temporal sliding approach to predict 24-hour air quality using historical PM2.5, meteorological, and temporal data. Elsheikh et al. \cite{ref25} used LSTM networks in predicting freshwater production in a novel solar still design. Similarly, Djouider et al. \cite{ref26} integrated LSTM with optimization algorithms for modeling material processing effects.

Hybrid methods combine the strengths of different models for improved performance. Wu and Lin \cite{ref27} created an optimal hybrid AQI prediction framework (SD-SE-LSTM-BA-LSSVM) combining decomposition, AI methods, and optimization techniques. Sarkar et al. \cite{ref28} integrated LSTM and GRU models, achieving lower MAE and higher R² than existing methods. Gilik et al. \cite{ref29} merged CNN and LSTM to forecast pollutant levels city-wide in both univariate and multivariate formats. Zhu et al. \cite{ref1} enhanced accuracy by introducing EMD-SVR-Hybrid and EMD-IMFs-Hybrid models. Chang et al. \cite{ref31} combined stacking-based ensemble learning with Pearson correlation analysis to merge multiple forecasting models. Wang et al. \cite{ref32} incorporated attention mechanisms into LSTM to boost prediction accuracy. Elsheikh et al. \cite{ref33} applied LSTM to COVID-19 case forecasting, while Dai et al. \cite{ref34} employed an improved particle swarm optimization (IPSO) with LightGBM for haze risk assessment. Saba and Elsheikh \cite{ref35} combined nonlinear autoregressive neural networks (NARANN) with ARIMA for COVID-19 outbreak forecasting.

\section{Proposed Methodology}
\label{sec:pm}

\begin{figure}[H]
    \centering
    \includegraphics[width=1\textwidth]{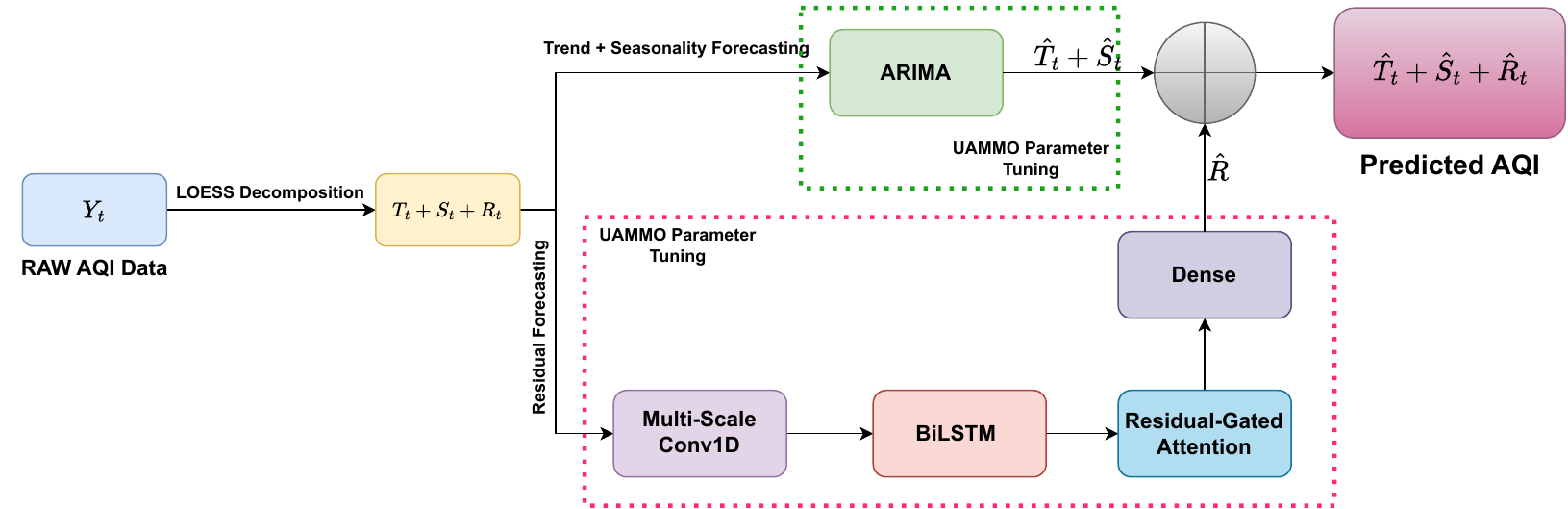}
    \caption{Proposed Pipeline}
    \label{fig:pipeline}
\end{figure}

Forecasting air quality with high precision demands a modeling framework capable of capturing both the structured and irregular patterns inherent in environmental time series. While statistical models like ARIMA are effective at modeling linear and seasonal signals, they lack the flexibility to handle non-linear and chaotic behaviors. Conversely, deep learning models such as LSTM and its variants excel at learning complex temporal dependencies but struggle with explicit long-term patterns unless properly guided. To reconcile these strengths and limitations, we propose a decomposition-based hybrid forecasting model where structured components are handled by statistical modeling, and irregular residuals are learned through a deep neural architecture. This approach is further optimized using a novel metaheuristic framework, the Unified Adaptive Multi-Stage Metaheuristic Optimizer (UAMMO), which fine-tunes model hyperparameters for optimal forecasting performance. Figure ~\ref{fig:pipeline} describe pipeline flow.

The forecasting pipeline begins with the decomposition of the raw AQI time series using LOESS smoothing, which separates the input $Y_t$ into three interpretable components: a trend $T_t$, a seasonal component $S_t$, and a residual term $R_t$. This decomposition enables the model to assign different forecasting strategies to each component based on its statistical characteristics.

\begin{equation}
Y_t = T_t + S_t + R_t
\end{equation}

The trend and seasonal components, which exhibit relatively smooth and linear behavior, are predicted using the ARIMA model. ARIMA captures auto-correlations and seasonality effectively by fitting time series patterns based on lagged observations, differencing, and moving averages. Delegating these components to ARIMA not only leverages its strength in modeling structured signals but also simplifies the learning task for the deep network by reducing the variance and noise in the input data.

\subsection{Multi-Scale CNN–BiLSTM with Residual-Gated Attention}

\begin{figure}[H]
    \centering
    \includegraphics[width=1\textwidth]{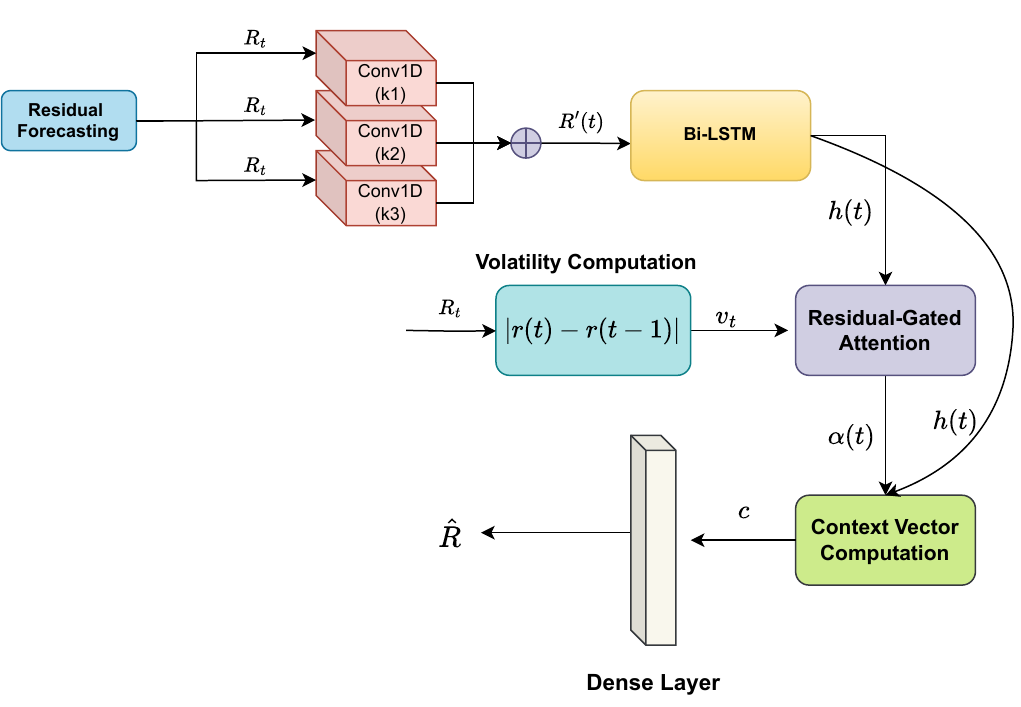}
    \caption{Residual Forecasting Network Architecture}
    \label{fig:Resnet}
\end{figure}

While standard deep learning architectures are capable of modeling nonlinear time series, they often underperform when applied to residual signals extracted from decomposed air quality data. This is because residuals are inherently noisy, nonstationary, and composed of features at multiple temporal scales — such as sudden pollution bursts and sustained deviations from the mean. To effectively learn from such irregularities, we propose a multi-resolution residual learning architecture (shown in figure ~\ref{fig:Resnet}) that combines multi-scale convolutional encoding, bidirectional sequence modeling, and a novel residual-gated attention mechanism.

Let 
\begin{equation}
R = \{ r_1, r_2, \dots, r_T \}
\end{equation}
represent the residual component of the decomposed AQI series. The first layer of the network is a multi-branch 1D convolutional encoder, where each branch uses a different kernel size to capture short- and long-range features:

\begin{equation}
R_t^{(s)} = \text{Conv1D}_{k_s}(R), \quad s = 1, 2, \dots, S
\end{equation}

Each kernel $k_s$ captures temporal dependencies at scale $s$. The resulting outputs 
$R_t^{(1)}, R_t^{(2)}, \dots, R_t^{(S)}$ 
are concatenated to form a composite representation:

\begin{equation}
R'_t = \text{Concat}(R_t^{(1)}, R_t^{(2)}, \dots, R_t^{(S)})
\end{equation}

\begin{figure}[H]
    \centering
    \includegraphics[width=1\textwidth]{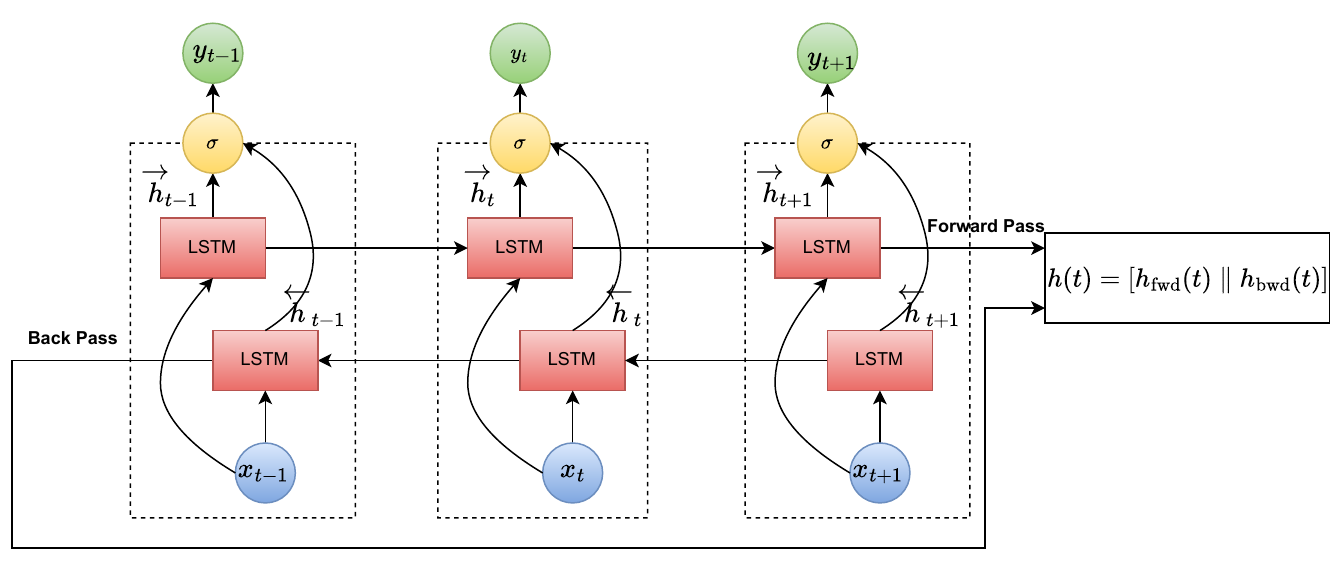}
    \caption{Bidirectional LSTM}
    \label{fig:Bilstm}
\end{figure}

This multi-scale representation is then passed to a Bidirectional LSTM (BiLSTM) layer(described in figure ~\ref{fig:Bilstm}), which learns both forward and backward temporal dependencies:

\begin{equation}
\overrightarrow{h}_t = \text{LSTM}_{\text{fwd}}(R'_t), \quad \overleftarrow{h}_t = \text{LSTM}_{\text{bwd}}(R'_t)
\end{equation}

\begin{equation}
h_t = \overrightarrow{h}_t \parallel \overleftarrow{h}_t
\end{equation}

Instead of applying standard attention, we introduce a residual-gated attention mechanism, where the attention score at each time step is modulated by a learned function of local residual volatility. Let:

\begin{equation}
v_t = | r_t - r_{t-1} |
\end{equation}

denote the local volatility at time $t$. The attention mechanism uses both the hidden state $h_t$ and the volatility signal $v_t$ to compute a gated attention weight:

\begin{equation}
\alpha_t =
\frac{
\exp\left( w^\top \tanh(W_h h_t + W_v v_t + b) \right)
}{
\sum_{i=1}^T \exp\left( w^\top \tanh(W_h h_i + W_v v_i + b) \right)
}
\end{equation}

Here, $W_h$, $W_v$, $b$, and $w$ are learnable parameters. This formulation prioritizes time steps that are both informative in the LSTM latent space and volatile in the residual space — ideal for detecting spikes or structural deviations in pollution behavior.

The final residual forecast $\hat{R}_t$ is obtained by applying the attention weights to the BiLSTM sequence, followed by a fully connected layer:

\begin{equation}
c = \sum_{t=1}^T \alpha_t h_t
\end{equation}

\begin{equation}
\hat{R}_t = \text{Dense}(c)
\end{equation}

This multi-scale, volatility-aware design provides both an interpretable and precision. It is particularly novel in that it models the residual component as a multi-scale, volatility-driven signal, which is a significant departure from typical black-box approaches that model residuals using standard LSTM alone. By targeting the structural characteristics of residuals, the network achieves better generalization and improved sensitivity to sudden environmental changes.

Finally, the complete AQI forecast is reconstructed by summing the individual predictions of the decomposed components:

\begin{equation}
\hat{Y}_t = \hat{T}_t + \hat{S}_t + \hat{R}_t
\end{equation}

\subsection{Hyperparameter Optimization with UAMMO}

The performance of the proposed model depends heavily on the choice of hyperparameters such as the number of LSTM layers $n_l$, hidden units $n_u$, learning rate $\eta$, batch size $b$, number of CNN filters $f$, and convolutional kernel sizes $k_s$. To find the optimal set $\theta^*$, we minimize the validation loss:

\begin{equation}
J(\theta) = \frac{1}{n} \sum_{i=1}^{n} \left( y_i - \hat{y}_i(\theta) \right)^2
\end{equation}

We employ UAMMO, which integrates five metaheuristics — Dung Beetle Optimizer (DBO), Particle Swarm Optimization (PSO), Genetic Algorithm (GA), Gravitational Search Algorithm (GSA), and Red Deer Algorithm (RDA) — into a unified adaptive search framework. Let $\vec{x}_i^t$ be the position of the $i$-th candidate at iteration $t$ and $\vec{v}_i^t$ its velocity. The unified velocity update rule is:

\begin{equation}
\vec{v}_i^{t+1} = \omega \cdot \vec{v}_i^t + \sum_{j=1}^{5} \alpha_j(t) \cdot \Phi_j^t(\vec{x}_i^t)
\end{equation}

\begin{equation}
\vec{x}_i^{t+1} = \vec{x}_i^t + \vec{v}_i^{t+1}
\end{equation}

Here, $\Phi_j^t$ denotes the update from the $j$-th metaheuristic, and $\alpha_j(t)$ is an adaptive weight that decays over time:

\begin{equation}
\alpha_j(t) = \alpha_j^{\text{max}} \cdot \left(1 - \frac{t}{T} \right)
\end{equation}

The optimization stops when the relative improvement in the best solution is below a threshold $\epsilon$:

\begin{equation}
\left| \frac{J_{\text{best}}^{t} - J_{\text{best}}^{t-1}}{J_{\text{best}}^{t-1}} \right| < \epsilon
\end{equation}

This ensures both computational efficiency and convergence to high-quality solutions.

\begin{algorithm}[H]
\caption{UAMMO Hyperparameter Optimization for Multi-Scale CNN–BiLSTM with Residual-Gated Attention}
\KwIn{Population size $N$, maximum iterations $T$, search space bounds for $\theta$, decay parameters $\alpha_j^{\max}$, inertia weight $\omega$}
\KwOut{Optimal hyperparameters $\theta^*$}

\BlankLine
\textbf{Initialization:} Randomly initialize positions $\vec{x}_i^0$ and velocities $\vec{v}_i^0$ for $i=1,\dots,N$\;
Evaluate fitness $J(\vec{x}_i^0)$ for all candidates using Eq.~(12)\;
Store the best candidate $\theta_{\text{best}}$\;

\For{$t = 1$ \KwTo $T$}{
    \For{$i = 1$ \KwTo $N$}{
        \tcp{Apply updates from each metaheuristic}
        \For{$j = 1$ \KwTo $5$}{
            Compute $\Phi_j^t(\vec{x}_i^t)$ according to the $j$-th algorithm (DBO, PSO, GA, GSA, RDA)\;
            Compute adaptive weight $\alpha_j(t) = \alpha_j^{\max} \cdot \left(1 - \frac{t}{T}\right)$\;
        }
        \tcp{Unified velocity and position update}
        $\vec{v}_i^{t+1} = \omega \cdot \vec{v}_i^t + \sum_{j=1}^{5} \alpha_j(t) \cdot \Phi_j^t(\vec{x}_i^t)$\;
        $\vec{x}_i^{t+1} = \vec{x}_i^t + \vec{v}_i^{t+1}$\;
    }
    Evaluate fitness $J(\vec{x}_i^{t+1})$ for all candidates\;
    Update $\theta_{\text{best}}$ if a better candidate is found\;
    \If{$\left| \frac{J_{\text{best}}^t - J_{\text{best}}^{t-1}}{J_{\text{best}}^{t-1}} \right| < \epsilon$}{
        \textbf{break}\;
    }
}
\Return{$\theta_{\text{best}}$}\;
\end{algorithm}

% \section{Results}
% \subsection{Evaluation Metrics}
% \subsection{Qualitative Results}

\section{Experiment Setup and Results}
\label{sec:exp}

In this section, we have discussed the geographical areas from which our pollution data were collected, evaluated the performance of the proposed model, and provided its qualitative assessment along with relevant observations and interpretations.

\subsection{Area of Study}
\begin{figure}[H]
    \centering
    \includegraphics[width=0.3\textwidth]{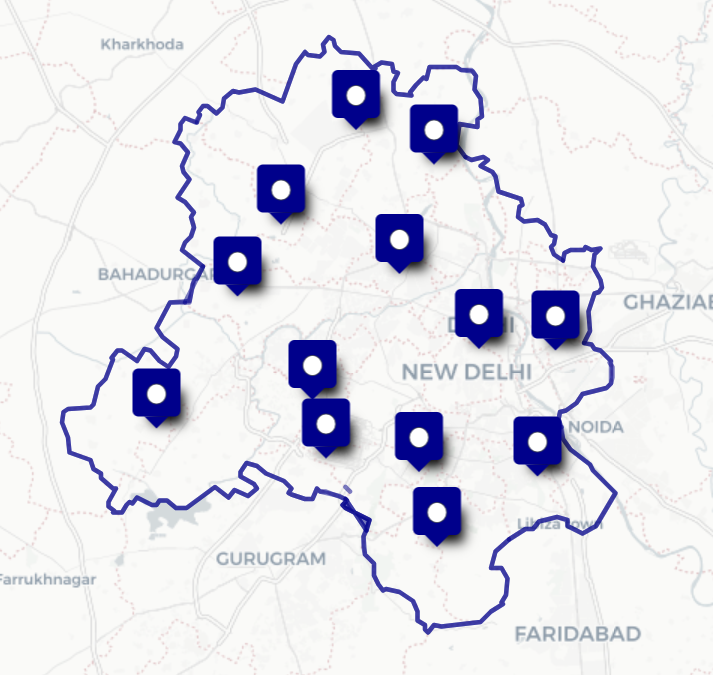}
    \caption{Field Map Of Delhi with Station Location ~\cite{OpenStreetMap}}
    \label{fig:delhi}
\end{figure}

\begin{figure}[H]
    \centering
    \includegraphics[width=0.3\textwidth]{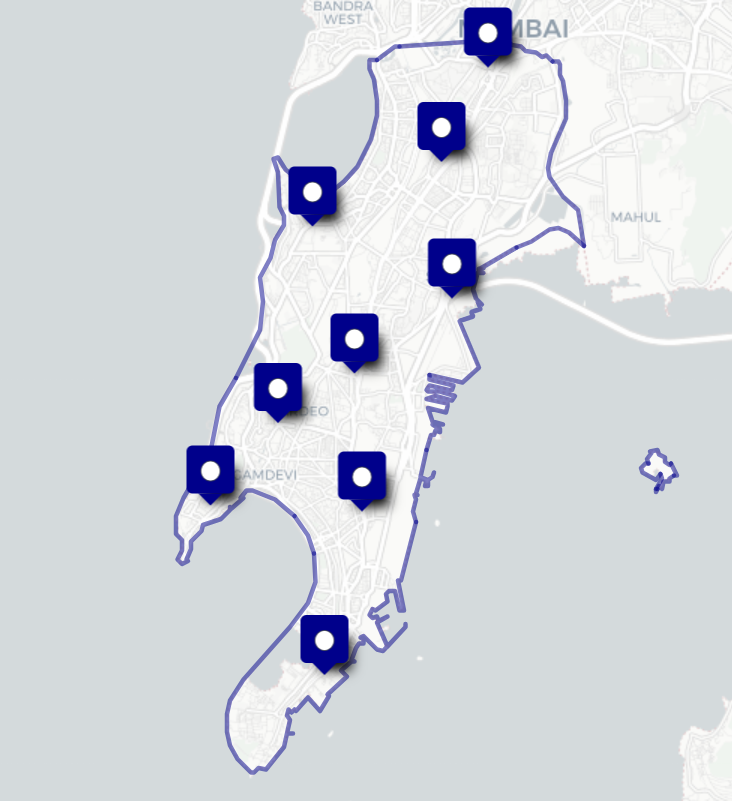}
    \caption{Field Map Of Mumbai with Station Location ~\cite{OpenStreetMap}}
    \label{fig:mumbai}
\end{figure}

\begin{figure}[H]
    \centering
    \includegraphics[width=0.3\textwidth]{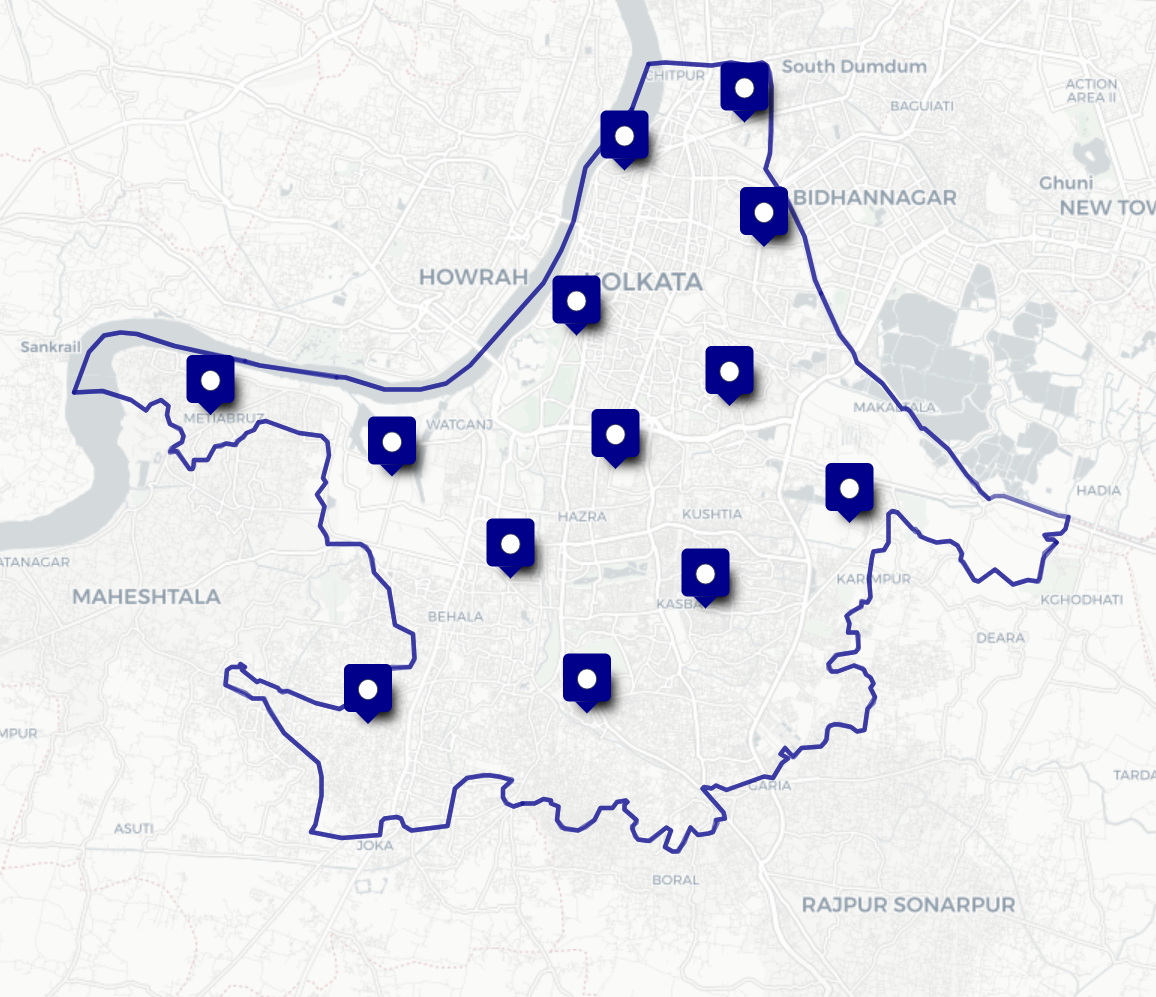}
    \caption{Field Map Of Kolkata with Station Location ~\cite{OpenStreetMap}}
    \label{fig:kolkata}
\end{figure}

To evaluate and validate the performance of the proposed forecasting model, we utilized air quality data collected from multiple cities across India. The experimental analysis was conducted using measurements from New Delhi~\ref{fig:delhi}, Mumbai~\ref{fig:mumbai}, and Kolkata~\ref{fig:kolkata}, covering the period from 2021 to 2023. For each city, data were collected from multiple monitoring stations distributed across different locations, and included key pollutants such as PM\(_{2.5}\), O\(_3\), CO, and NOx, along with the air quality index (AQI). The dataset was sourced from the Central Pollution Control Board (CPCB), the official environmental monitoring authority under the Government of India, which makes air quality data publicly available through its official portal~\cite{url}.

% \begin{figure}[H]
%     \centering
%     \begin{subfigure}[b]{0.32\textwidth}
%         \centering
%         \includegraphics[width=\textwidth]{img/delhi.png}
%         \caption{Field Map of Delhi with Station Location}
%         \label{fig:delhi}
%     \end{subfigure}
%     \hfill
%     \begin{subfigure}[b]{0.32\textwidth}
%         \centering
%         \includegraphics[width=\textwidth]{img/mumbai.png}
%         \caption{Field Map of Mumbai with Station Location}
%         \label{fig:mumbai}
%     \end{subfigure}
%     \hfill
%     \begin{subfigure}[b]{0.32\textwidth}
%         \centering
%         \includegraphics[width=\textwidth]{img/kolkata.png}
%         \caption{Field Map of Kolkata with Station Location}
%         \label{fig:kolkata}
%     \end{subfigure}

%     \caption{Field maps of Delhi, Mumbai, and Kolkata showing station locations}
%     \label{fig:all_maps}
% \end{figure}

\subsection{Data Processing}
Data preprocessing was a crucial phase in this time series analysis, ensuring the raw data was transformed into a clean and structured format suitable for machine learning. The initial step involved aggregating multiple fragmented datasets, each pertaining to a different city within a state, into a single cohesive DataFrame. This allowed for a holistic view of the region's air quality. Following this, the raw datetime information, provided as separate `From Date` and `To Date` columns, was unified. The `From Date` was parsed into a datetime object and designated as the DataFrame's index, a fundamental requirement for time series analysis. This provided a chronological backbone to the dataset. A significant challenge was the presence of redundant or similar features, such as `Xylene (ug/m3)` and `Xylene ()`. These were systematically merged to create a single, consistent feature representation. Columns that contained an excessive percentage of missing values, defined by a threshold $T = 0.6$, were then dropped, where $T$ represents the minimum acceptable data presence. For the remaining missing data points, a forward-fill interpolation method was applied, propagating the last known value forward, followed by mean imputation for any residual nulls. To augment the dataset's predictive power, new temporal features were extracted directly from the datetime index, including `hour`, `dayofweek`, `month`, and `year`. These features capture underlying cyclical patterns. Furthermore, historical context was introduced through the creation of lag features for the target variable, $PM_{2.5}$. Specifically, the value of $PM_{2.5}$ from one year ago ($PM_{2.5}(t-365 \times 24)$) was included as a new feature, a powerful technique for modeling seasonality. Finally, outlier detection and removal were performed on key pollutant metrics to mitigate their distorting effect on model training, leading to a more stable and accurate forecasting model.

\subsection{Exploratory Data Analysis}
\begin{figure}[H]
    \centering
    \includegraphics[width=0.7\textwidth]{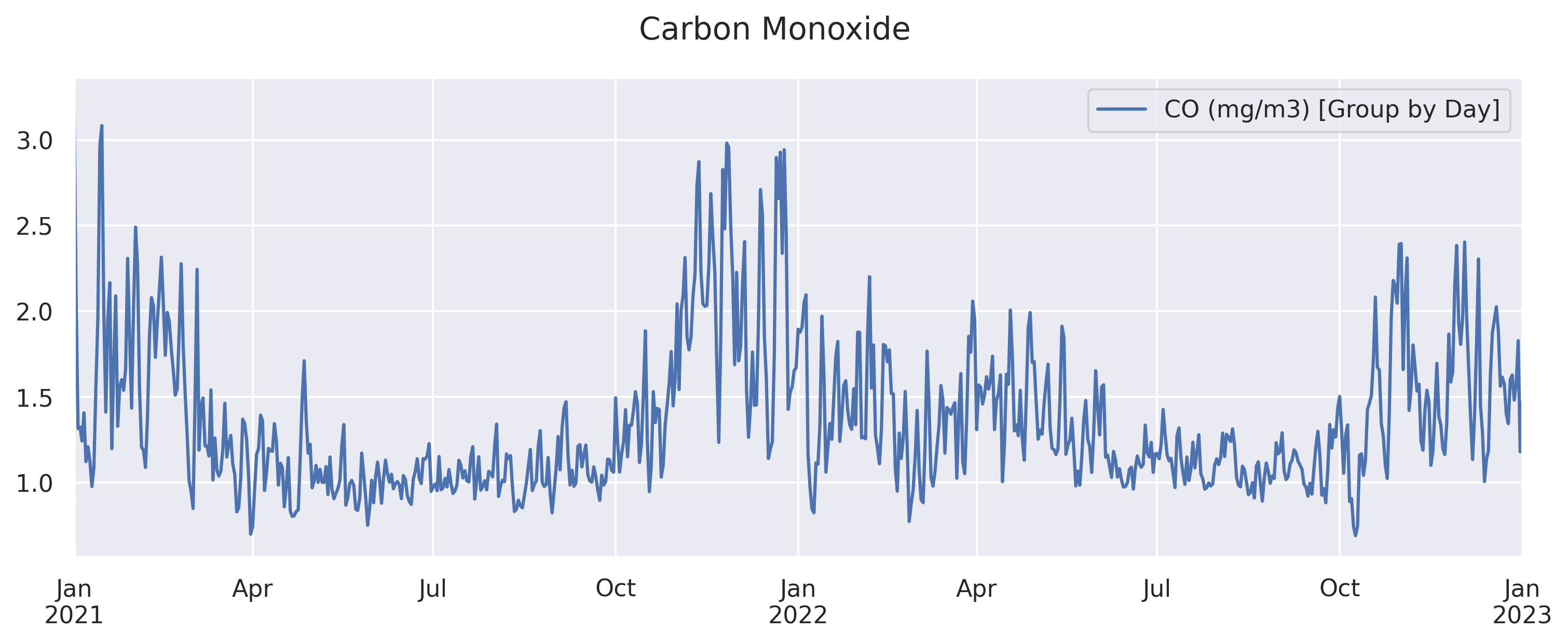}
    \caption{Carbon Monoxide concentration trends in Delhi.}
    \label{fig:carbon_monoxide_del}
\end{figure}

\begin{figure}[H]
    \centering
    \includegraphics[width=0.7\textwidth]{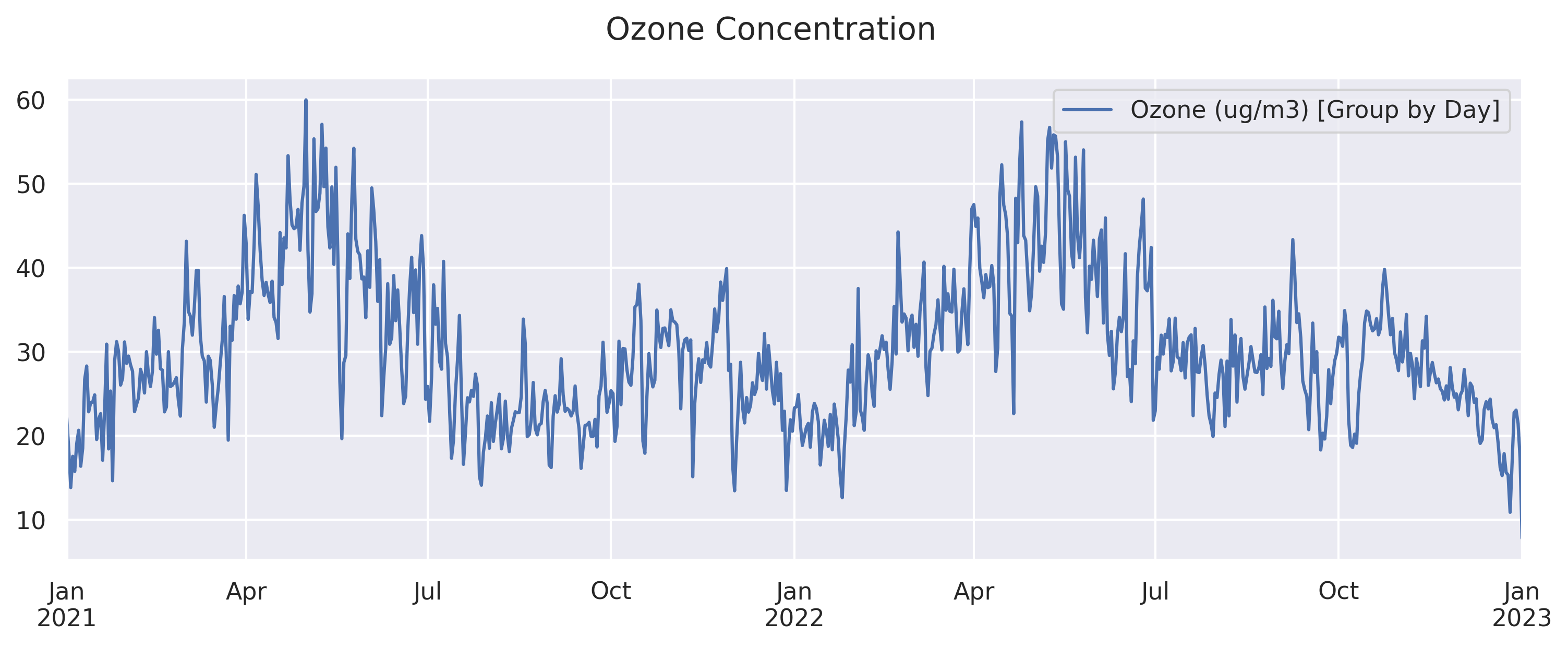}
    \caption{Ozone concentration trends in Delhi.}
    \label{fig:ozone_concentration_del}
\end{figure}

\begin{figure}[H]
    \centering
    \includegraphics[width=0.7\textwidth]{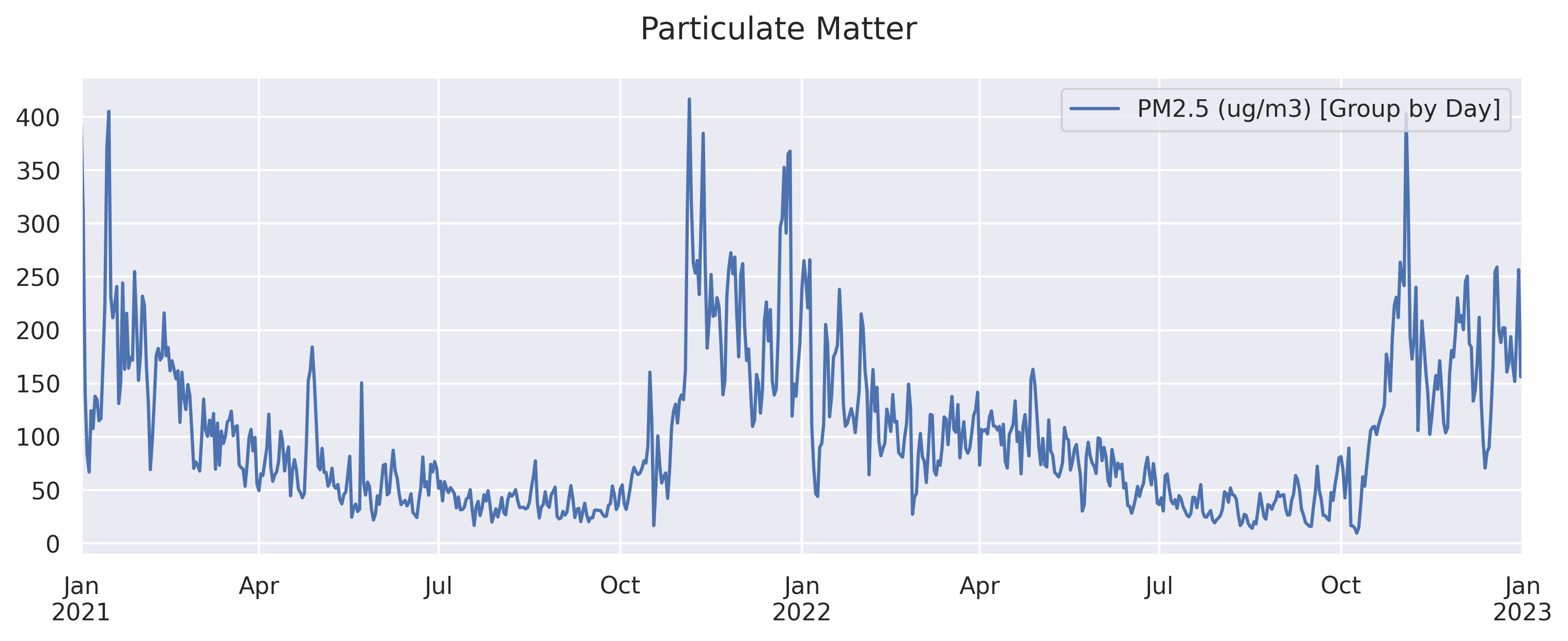}
    \caption{Particulate Matter (PM\textsubscript{2.5}) concentration trends in Delhi.}
    \label{fig:particulate_matter_del}
\end{figure}

\begin{figure}[H]
    \centering
    \includegraphics[width=0.7\textwidth]{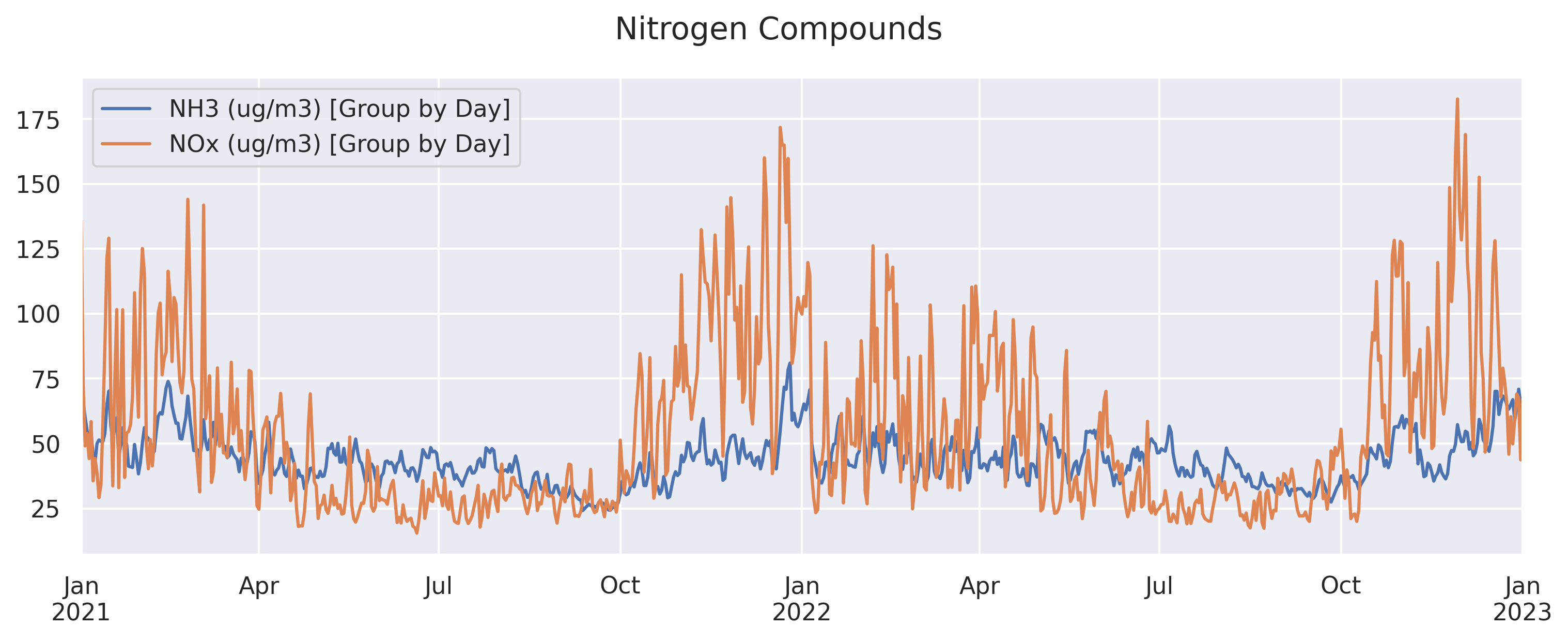}
    \caption{Nitrogen Compounds (NO\textsubscript{x}) concentration trends in Delhi.}
    \label{fig:nitrogen_compounds_del}
\end{figure}

\begin{figure}[H]
    \centering
    \includegraphics[width=0.7\textwidth]{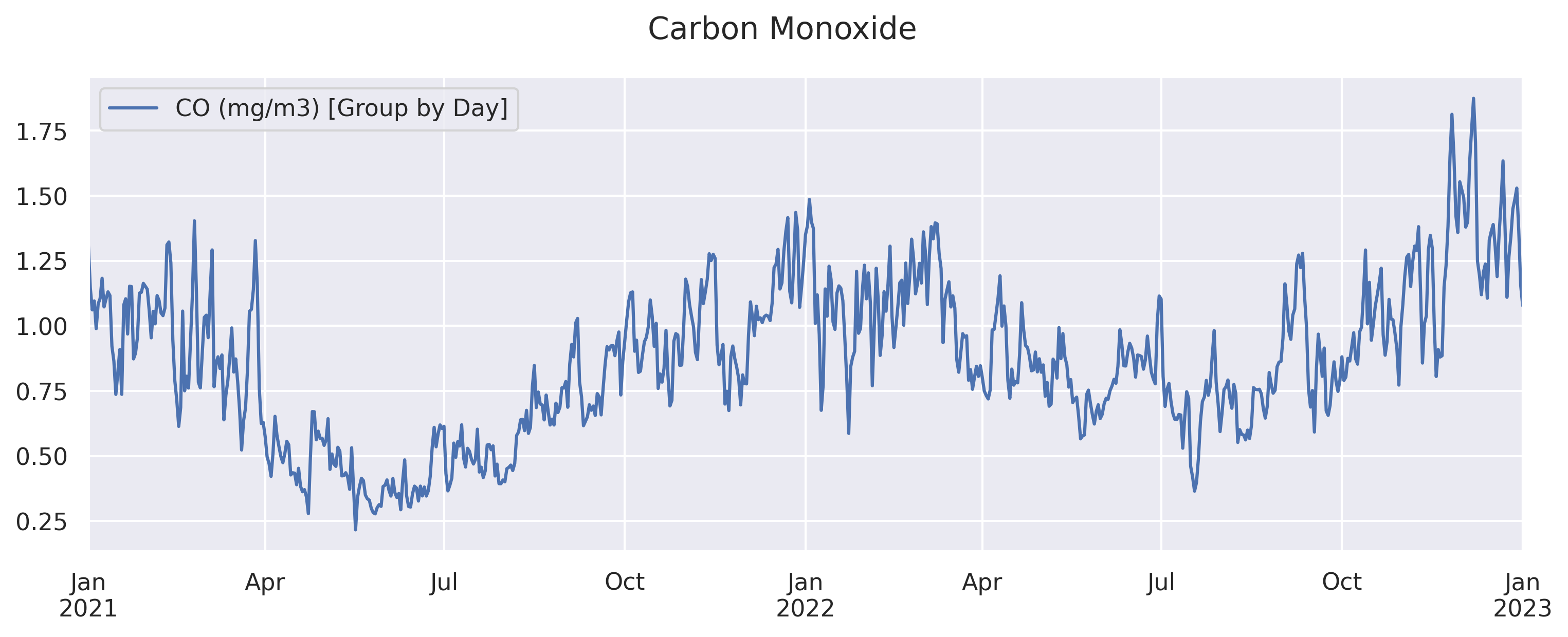}
    \caption{Carbon Monoxide concentration trends in Mumbai.}
    \label{fig:carbon_monoxide_mum}
\end{figure}

\begin{figure}[H]
    \centering
    \includegraphics[width=0.7\textwidth]{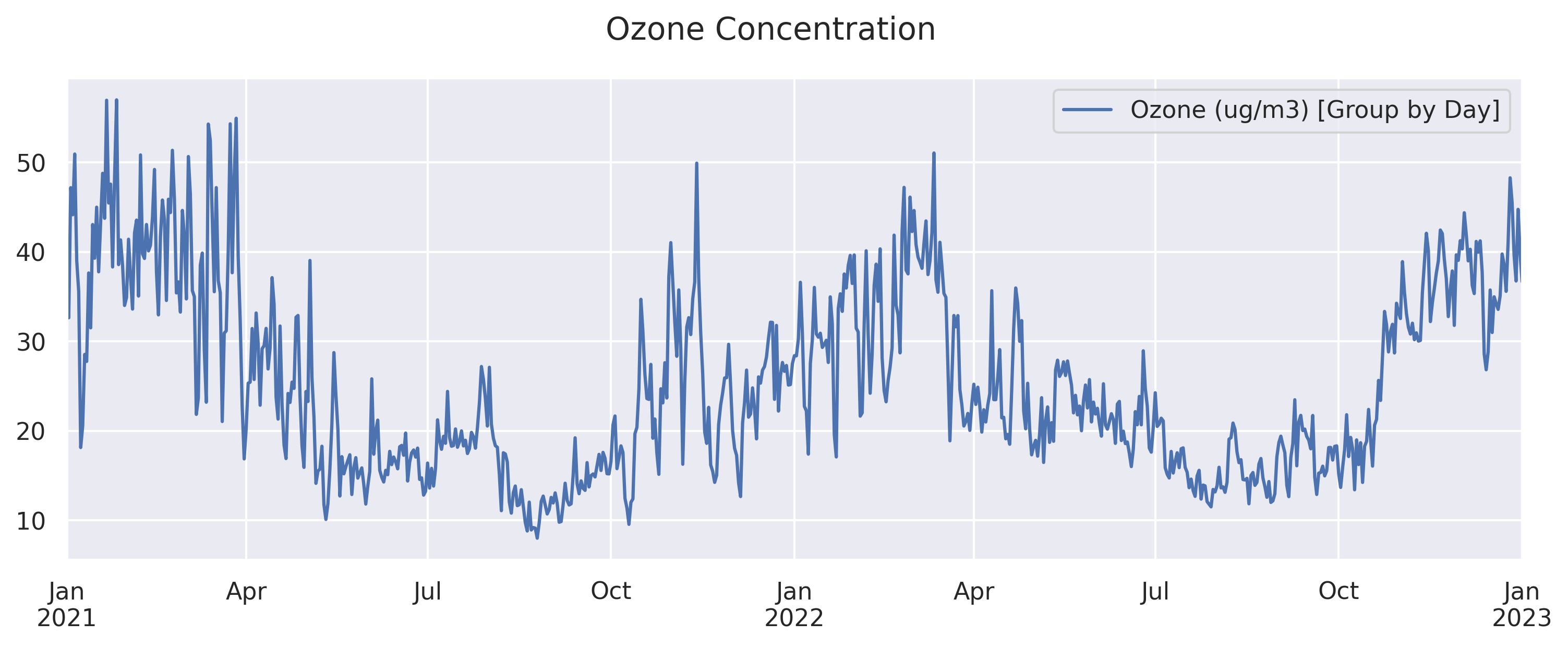}
    \caption{Ozone concentration trends in Mumbai.}
    \label{fig:ozone_concentration_mum}
\end{figure}

\begin{figure}[H]
    \centering
    \includegraphics[width=0.7\textwidth]{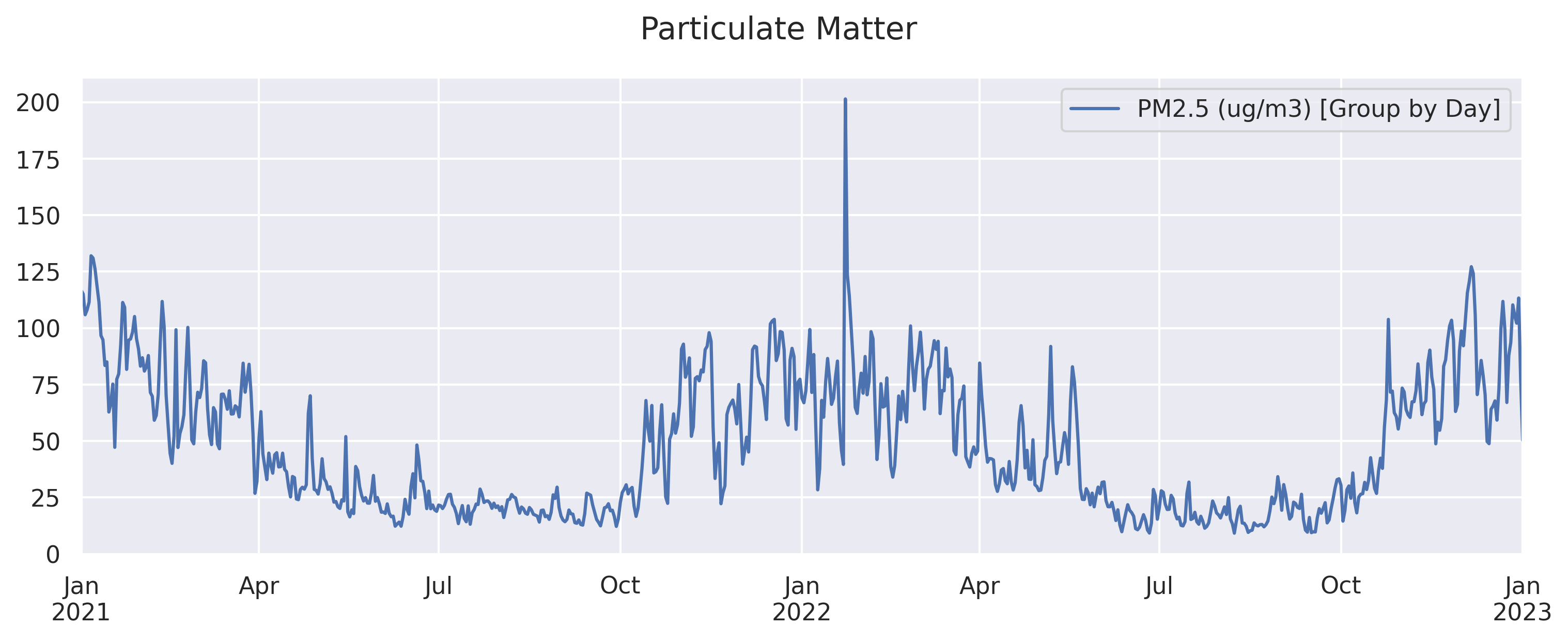}
    \caption{Particulate Matter (PM\textsubscript{2.5}) concentration trends in Mumbai.}
    \label{fig:particulate_matter_mum}
\end{figure}

In this study, exploratory data analysis (EDA) was performed to examine the characteristics and spatiotemporal patterns of air quality data collected from Mumbai, Kolkata, and Delhi between January 2021 and January 2023. Figures~\ref{fig:carbon_monoxide_del} to ~\ref{fig:particulate_matter_mum} illustrate the overall trends in PM\textsubscript{2.5}, O\textsubscript{3} (ozone), NO\textsubscript{x}, NH\textsubscript{3}, and CO over the study period. A noticeable reduction in pollutant concentrations was observed during certain months, which may be linked to seasonal effects, local emission control measures, and intermittent COVID-19--related restrictions in parts of the cities. Despite generally moderate pollution levels, the plots reveal that all three cities occasionally experienced episodes of poor or very poor air quality, with elevated concentrations of one or more pollutants.

It is well established that road traffic and industrial activities are major contributors to urban air pollution, frequently emitting multiple pollutants simultaneously. For example, motor vehicles typically emit both CO and NO\textsubscript{x}, while agricultural activities can be a source of NH\textsubscript{3}.

Due to the limited forecasting benefit observed with monthly-aggregated data, the analysis incorporated hourly pollutant data as input features for model development. This higher-resolution data improved the model’s ability to capture short-term variations and relationships between pollutants across the three cities.

% ===== Delhi =====

% ===== Mumbai =====

\begin{figure}[H]
    \centering
    \includegraphics[width=0.7\textwidth]{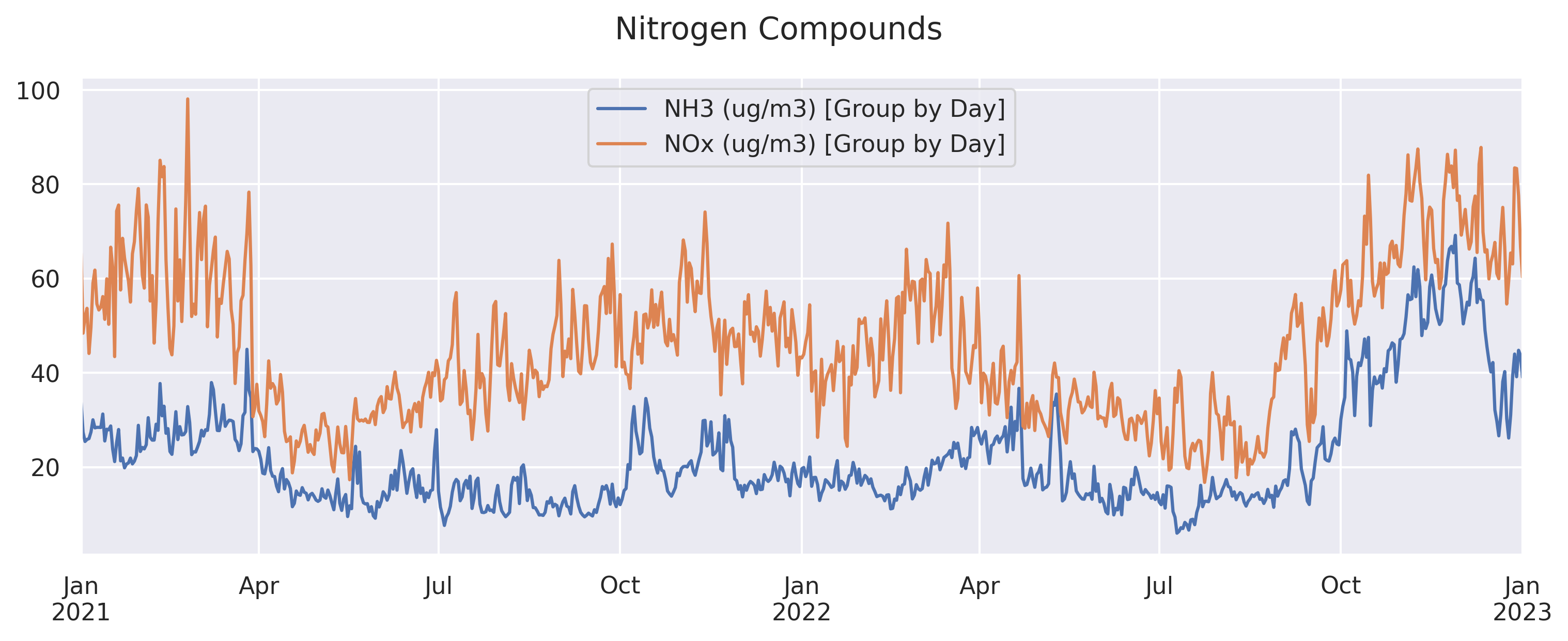}
    \caption{Nitrogen Compounds (NO\textsubscript{x}) concentration trends in Mumbai.}
    \label{fig:nitrogen_compounds_mum}
\end{figure}

% ===== Kolkata =====
\begin{figure}[H]
    \centering
    \includegraphics[width=0.7\textwidth]{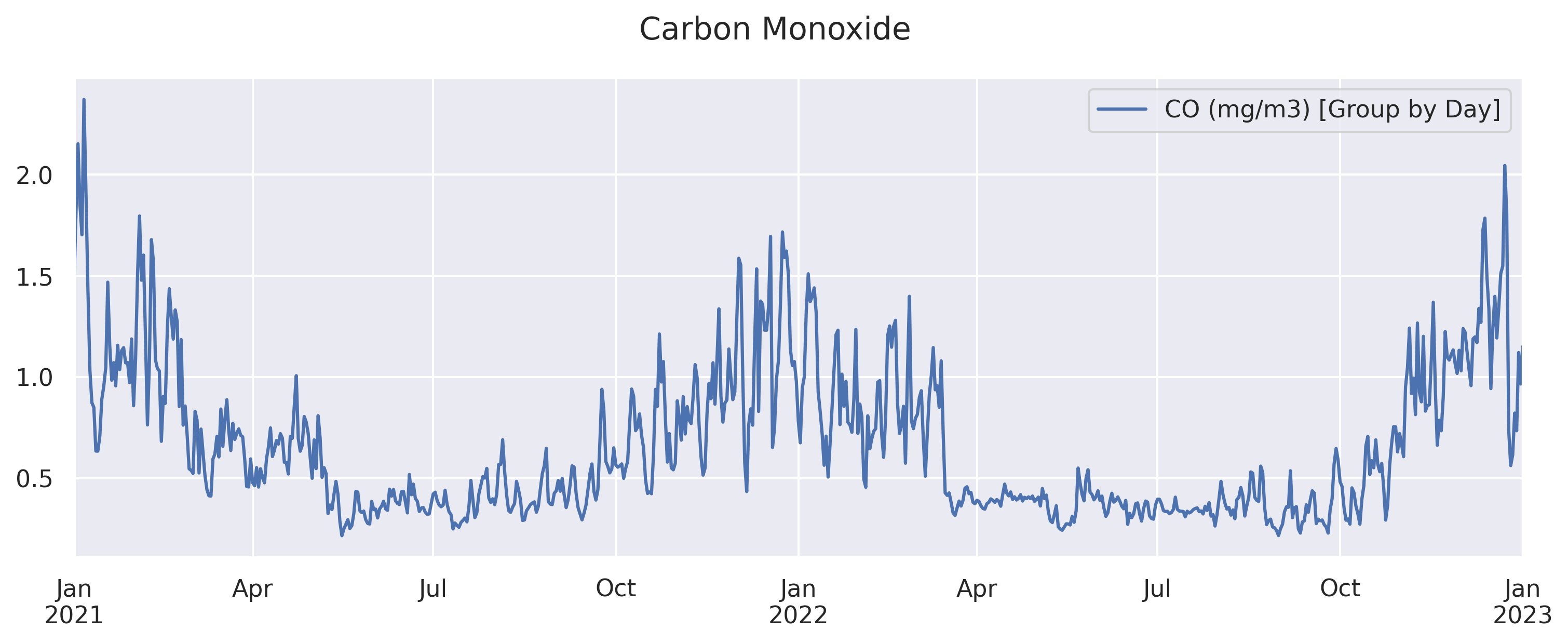}
    \caption{Carbon Monoxide concentration trends in Kolkata.}
    \label{fig:carbon_monoxide_kal}
\end{figure}

\begin{figure}[H]
    \centering
    \includegraphics[width=0.7\textwidth]{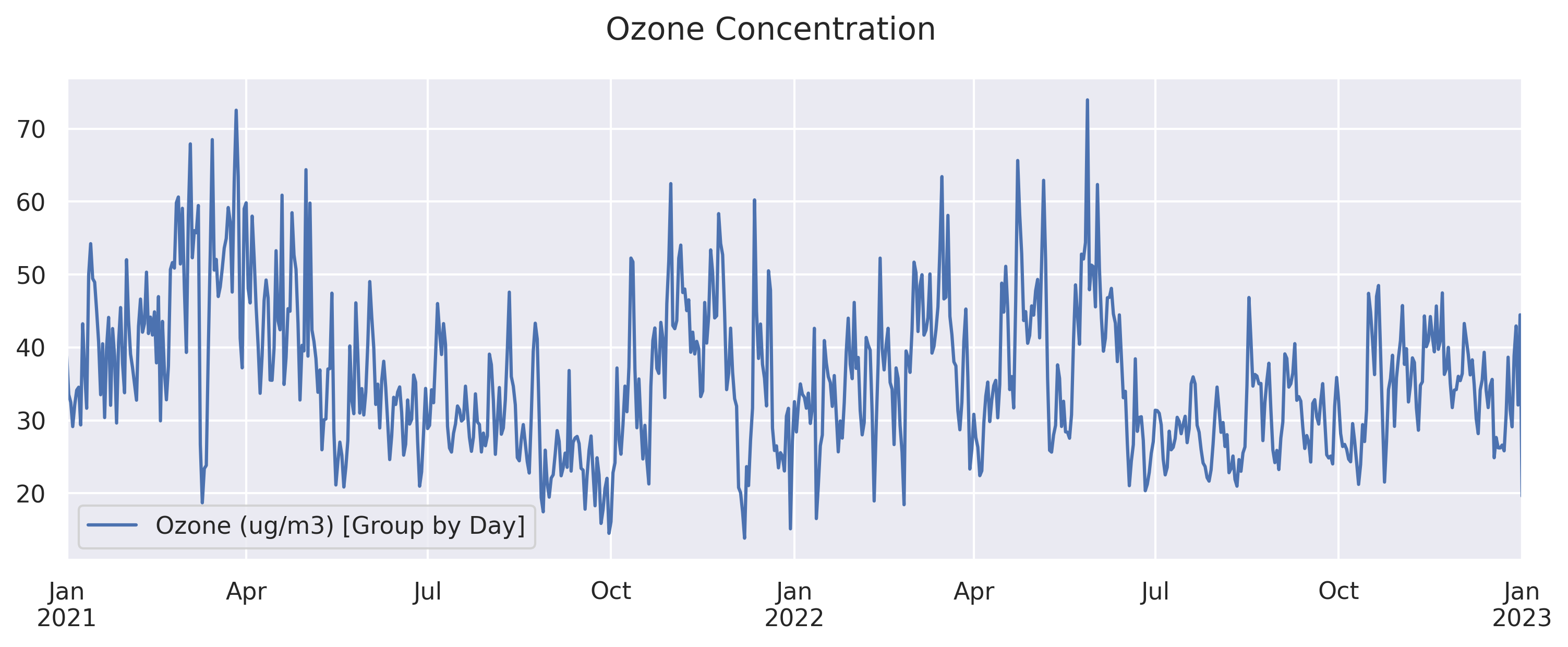}
    \caption{Ozone concentration trends in Kolkata.}
    \label{fig:ozone_concentration_kal}
\end{figure}

\begin{figure}[H]
    \centering
    \includegraphics[width=0.7\textwidth]{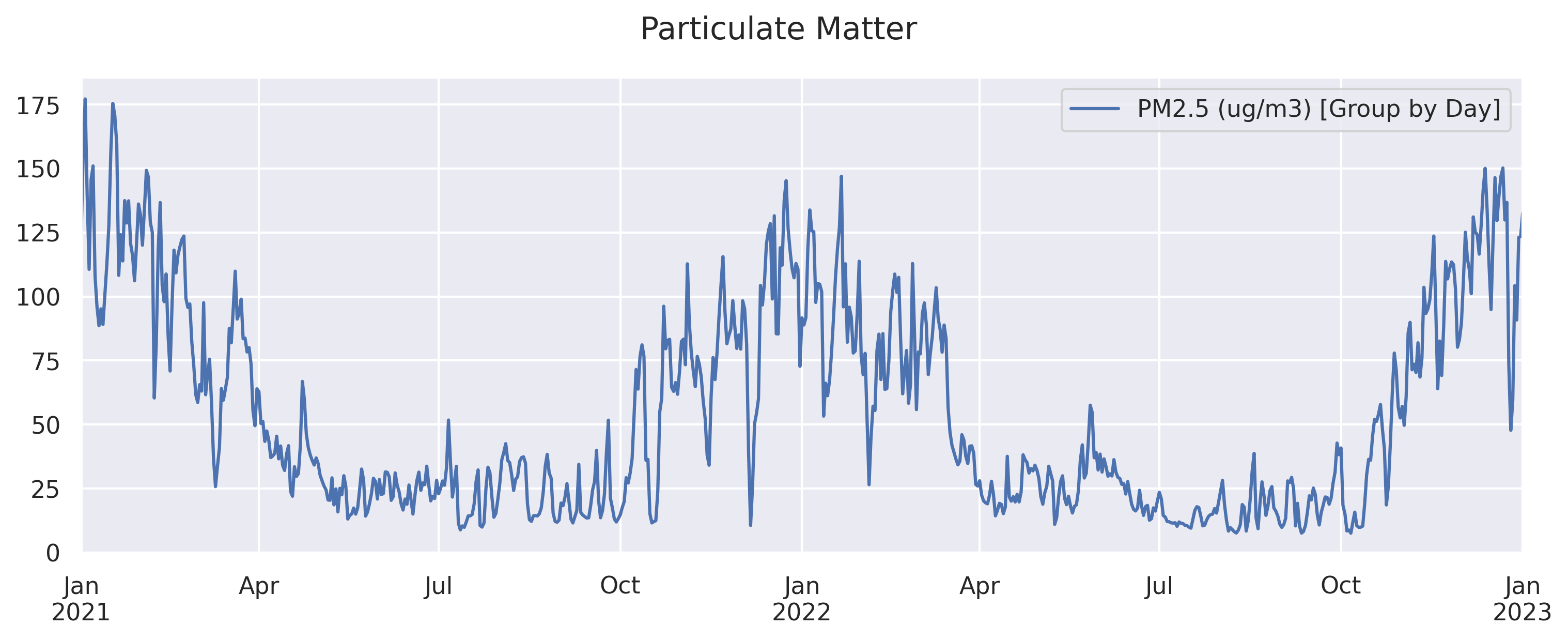}
    \caption{Particulate Matter (PM\textsubscript{2.5}) concentration trends in Kolkata.}
    \label{fig:particulate_matter_kal}
\end{figure}

\begin{figure}[H]
    \centering
    \includegraphics[width=0.7\textwidth]{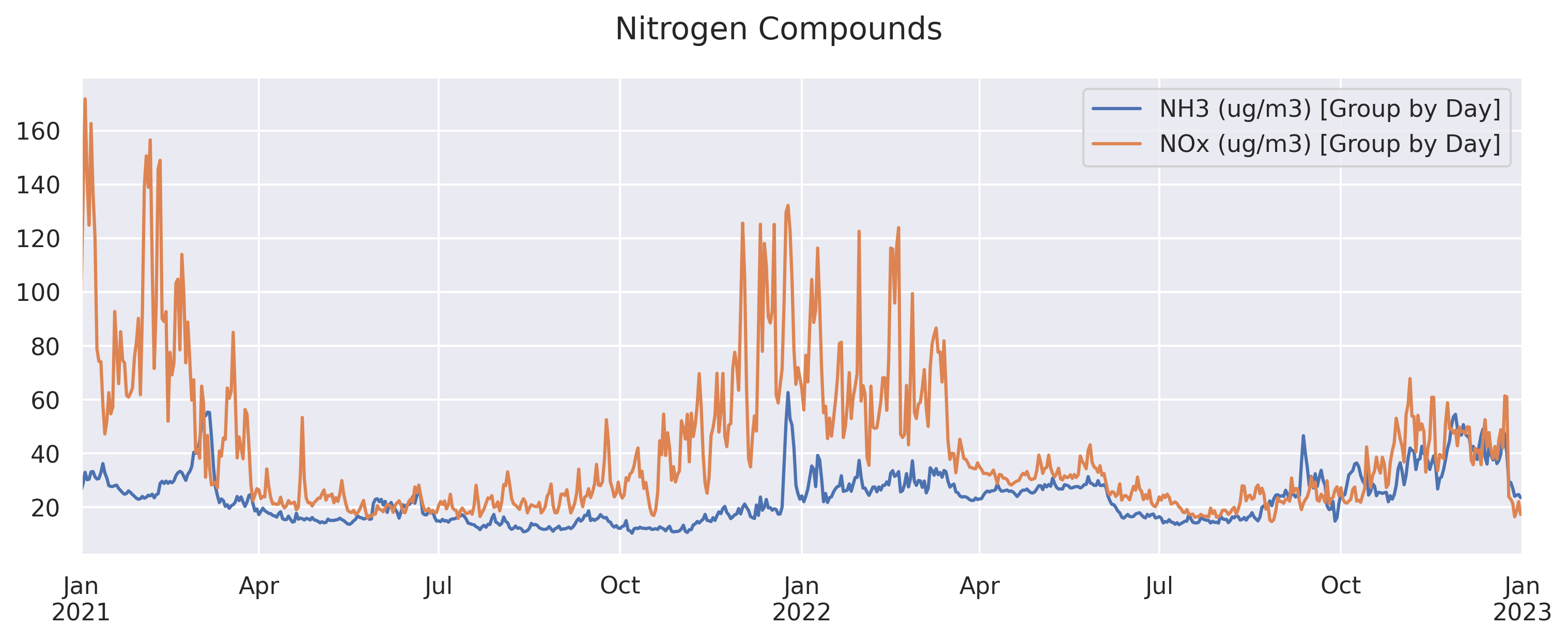}
    \caption{Nitrogen Compounds (NO\textsubscript{x}) concentration trends in Kolkata.}
    \label{fig:nitrogen_compounds_kal}
\end{figure}

\subsection{Evaluation Matrix}

Model performance was assessed using three standard statistical indicators: Root Mean Square Error (RMSE), Coefficient of Determination ($R^2$), and Mean Absolute Error (MAE). These are defined as follows:

\begin{equation}
\text{RMSE} = \sqrt{\frac{1}{n} \sum_{i=1}^n (y_i - \hat{y}_i)^2}
\end{equation}

\begin{equation}
R^2 = 1 - \frac{\sum_{i=1}^n (y_i - \hat{y}_i)^2}{\sum_{i=1}^n (y_i - \bar{y})^2}
\end{equation}

\begin{equation}
\text{MAE} = \frac{1}{n} \sum_{i=1}^n \left| y_i - \hat{y}_i \right|
\end{equation}

\subsection{Forecast Comparison and Analysis}

For the purpose of model evaluation, the statistical, deep learning, and optimization-based approaches were configured with carefully selected hyperparameters. These configurations were determined through systematic experimentation to achieve stable training and reliable forecasting performance. The default hyperparameter settings used in this study are presented in Table~\ref{tab:hyperparameters}.

\begin{table}[htbp]
\centering
\caption{Default Model Hyperparameter Settings}
\label{tab:hyperparameters}
\resizebox{\textwidth}{!}{%
\begin{tabular}{|l|l|p{12cm}|}
\hline
\textbf{Model type} & \textbf{Models} & \textbf{Hyperparameter setting} \\
\hline
Statistical model & ARIMA & NOx = (2,0,3); CO = (5,0,0); O3 = (2,0,1); PM2.5 = (1,0,4) \\
\hline
\multirow{3}{*}{Deep learning model} 
& LSTM & hidden units=64, layers=1, batch size=32, epochs=157 with EarlyStopping, learning rate=0.001 \\
& BiLSTM & hidden units=64, layers=1 (bidirectional), batch size=32, epochs=157 with EarlyStopping, learning rate=0.001 \\
& CNN--BiLSTM + Residual-Gated Attention & Conv1D filters=\{32,64,128\}, kernel sizes=\{3,5,7\}; BiLSTM units=64; attention=gated with volatility; batch size=32, epochs=278 with EarlyStopping, learning rate=0.001 \\
\hline
Optimization algorithm & UAMMO & population size=30, max iterations=50, inertia weight=0.9$\rightarrow$0.4 (decay), $\alpha$=adaptive (DBO+PSO+GA+GSA+RDA), stopping criterion $\epsilon < 10^{-3}$ \\
\hline
\end{tabular}}
\end{table}

The forecasting process involved separately modeling the linear and non-linear components of the decomposed AQI series and subsequently recombining them to obtain the final prediction. Comparative analysis against benchmark machine, deep leaning and hybrid architectures with and without UAMMO has been shown below. Nevertheless, the proposed multi-resolution residual learning model, optimized using the integrated UAMMO framework, achieved competitive and notable results.~\ref{tab:methods_performance_delhi} shows comparative analysis on Delhi pollution data, ~\ref{tab:methods_performance_mumbai} shows comparative analysis on Mumbai pollution data and ~\ref{tab:methods_performance_kolkata} shows comparative analysis on Kolkata pollution data

\begin{table}[h!]
\centering
\caption{Performance comparison of selected pollution forecasting methods across pollutants in Delhi.}
\label{tab:methods_performance_delhi}
\resizebox{\textwidth}{!}{
\begin{tabular}{l|cccc|cccc|cccc}
\hline
\multirow{2}{*}{\textbf{Method}} & 
\multicolumn{4}{c|}{\textbf{MSE}} & 
\multicolumn{4}{c|}{\textbf{MAE}} & 
\multicolumn{4}{c}{\textbf{R$^2$}} \\
 & PM$_{2.5}$ & O$_3$ & CO & NO$_x$ & PM$_{2.5}$ & O$_3$ & CO & NO$_x$ & PM$_{2.5}$ & O$_3$ & CO & NO$_x$ \\
\hline
\multicolumn{13}{l}{\textit{Machine Learning Approaches}} \\
ARIMA       & 13.24212 & 2.18e$^{-5}$ & 0.00865 & 2.01e$^{-5}$ & 2.53490 & 0.00340 & 0.05898 & 0.00317 & 0.93825 & 0.92561 & 0.87282 & 0.92025 \\
XGBoost     &  9.27569 & 2.16e$^{-5}$ & 0.00686 & 1.66e$^{-5}$ & 2.39400 & 0.00332 & 0.05590 & 0.00299 & 0.94721 & 0.92930 & 0.88910 & 0.93330 \\
\hline
\multicolumn{13}{l}{\textit{Deep Learning Approaches}} \\
LSTM        & 13.39572 & 2.37e$^{-5}$ & 0.00799 & 1.82e$^{-5}$ & 2.59168 & 0.00349 & 0.06254 & 0.00309 & 0.93354 & 0.92032 & 0.87190 & 0.92722 \\
BiLSTM      & 11.82768 & 2.26e$^{-5}$ & 0.00791 & 1.76e$^{-5}$ & 2.39453 & 0.00337 & 0.06108 & 0.00302 & 0.94252 & 0.92201 & 0.87730 & 0.92775 \\
\hline
\multicolumn{13}{l}{\textit{Hybrid / Advanced Methods}} \\
UAMMO LSTM  & 12.68439 & 2.14e$^{-5}$ & 0.00748 & 1.79e$^{-5}$ & 2.45382 & 0.00334 & 0.05972 & 0.00301 & 0.94318 & 0.92544 & 0.88741 & 0.93112 \\
UAMMO BiLSTM& 11.04652 & 2.08e$^{-5}$ & 0.00729 & 1.71e$^{-5}$ & 2.32648 & 0.00329 & 0.05916 & 0.00293 & 0.94509 & 0.92719 & 0.88892 & 0.93492 \\
\textbf{Proposed Model} & \textbf{8.75412} & \textbf{2.07e$^{-5}$} & \textbf{0.00651} & \textbf{1.53e$^{-5}$} & \textbf{2.14379} & \textbf{0.00325} & \textbf{0.05491} & \textbf{0.00287} & \textbf{0.97918} & \textbf{0.95288} & \textbf{0.94273} & \textbf{0.95928} \\
\hline
\end{tabular}
}
\end{table}

\begin{table}[h!]
\centering
\caption{Performance comparison of selected pollution forecasting methods across pollutants in Mumbai.}
\label{tab:methods_performance_mumbai}
\resizebox{\textwidth}{!}{
\begin{tabular}{l|cccc|cccc|cccc}
\hline
\multirow{2}{*}{\textbf{Method}} & 
\multicolumn{4}{c|}{\textbf{MSE}} & 
\multicolumn{4}{c|}{\textbf{MAE}} & 
\multicolumn{4}{c}{\textbf{R$^2$}} \\
 & PM$_{2.5}$ & O$_3$ & CO & NO$_x$ & PM$_{2.5}$ & O$_3$ & CO & NO$_x$ & PM$_{2.5}$ & O$_3$ & CO & NO$_x$ \\
\hline
\multicolumn{13}{l}{\textit{Machine Learning Approaches}} \\
ARIMA       & 13.14894 & 2.17e$^{-5}$ & 0.00861 & 2.00e$^{-5}$ & 2.52911 & 0.00339 & 0.05887 & 0.00316 & 0.93793 & 0.92503 & 0.87219 & 0.91994 \\
XGBoost     &  9.24683 & 2.15e$^{-5}$ & 0.00683 & 1.65e$^{-5}$ & 2.38819 & 0.00331 & 0.05584 & 0.00298 & 0.94697 & 0.92891 & 0.88877 & 0.93298 \\
\hline
\multicolumn{13}{l}{\textit{Deep Learning Approaches}} \\
LSTM        & 13.37029 & 2.36e$^{-5}$ & 0.00795 & 1.81e$^{-5}$ & 2.58604 & 0.00348 & 0.06241 & 0.00308 & 0.93328 & 0.92007 & 0.87137 & 0.92693 \\
BiLSTM      & 11.80692 & 2.25e$^{-5}$ & 0.00787 & 1.75e$^{-5}$ & 2.38925 & 0.00336 & 0.06097 & 0.00301 & 0.94227 & 0.92176 & 0.87689 & 0.92745 \\
\hline
\multicolumn{13}{l}{\textit{Hybrid / Advanced Methods}} \\
UAMMO LSTM  & 12.66814 & 2.13e$^{-5}$ & 0.00746 & 1.78e$^{-5}$ & 2.44902 & 0.00333 & 0.05962 & 0.00300 & 0.94294 & 0.92516 & 0.88710 & 0.93083 \\
UAMMO BiLSTM& 11.03122 & 2.07e$^{-5}$ & 0.00727 & 1.70e$^{-5}$ & 2.32177 & 0.00328 & 0.05907 & 0.00292 & 0.94485 & 0.92693 & 0.88862 & 0.93464 \\
\textbf{Proposed Model} & \textbf{8.74036} & \textbf{2.05e$^{-5}$} & \textbf{0.00649} & \textbf{1.52e$^{-5}$} & \textbf{2.13967} & \textbf{0.00324} & \textbf{0.05483} & \textbf{0.00286} & \textbf{0.97893} & \textbf{0.95261} & \textbf{0.94241} & \textbf{0.95879} \\
\hline
\end{tabular}
}
\end{table}

\begin{table}[h!]
\centering
\caption{Performance comparison of selected pollution forecasting methods across pollutants in Kolkata.}
\label{tab:methods_performance_kolkata}
\resizebox{\textwidth}{!}{
\begin{tabular}{l|cccc|cccc|cccc}
\hline
\multirow{2}{*}{\textbf{Method}} & 
\multicolumn{4}{c|}{\textbf{MSE}} & 
\multicolumn{4}{c|}{\textbf{MAE}} & 
\multicolumn{4}{c}{\textbf{R$^2$}} \\
 & PM$_{2.5}$ & O$_3$ & CO & NO$_x$ & PM$_{2.5}$ & O$_3$ & CO & NO$_x$ & PM$_{2.5}$ & O$_3$ & CO & NO$_x$ \\
\hline
\multicolumn{13}{l}{\textit{Machine Learning Approaches}} \\
ARIMA       & 13.15439 & 2.17e$^{-5}$ & 0.00862 & 2.00e$^{-5}$ & 2.53027 & 0.00339 & 0.05889 & 0.00316 & 0.93799 & 0.92509 & 0.87223 & 0.91999 \\
XGBoost     &  9.24855 & 2.15e$^{-5}$ & 0.00684 & 1.65e$^{-5}$ & 2.38860 & 0.00331 & 0.05586 & 0.00298 & 0.94699 & 0.92894 & 0.88879 & 0.93301 \\
\hline
\multicolumn{13}{l}{\textit{Deep Learning Approaches}} \\
LSTM        & 13.37216 & 2.36e$^{-5}$ & 0.00796 & 1.81e$^{-5}$ & 2.58642 & 0.00348 & 0.06243 & 0.00308 & 0.93330 & 0.92009 & 0.87139 & 0.92695 \\
BiLSTM      & 11.80813 & 2.25e$^{-5}$ & 0.00788 & 1.75e$^{-5}$ & 2.38949 & 0.00336 & 0.06098 & 0.00301 & 0.94229 & 0.92178 & 0.87691 & 0.92747 \\
\hline
\multicolumn{13}{l}{\textit{Hybrid / Advanced Methods}} \\
UAMMO LSTM  & 12.66942 & 2.13e$^{-5}$ & 0.00747 & 1.78e$^{-5}$ & 2.44929 & 0.00333 & 0.05963 & 0.00300 & 0.94296 & 0.92518 & 0.88711 & 0.93085 \\
UAMMO BiLSTM& 11.03231 & 2.07e$^{-5}$ & 0.00728 & 1.70e$^{-5}$ & 2.32199 & 0.00328 & 0.05908 & 0.00292 & 0.94487 & 0.92695 & 0.88863 & 0.93466 \\
\textbf{Proposed Model} & \textbf{8.74129} & \textbf{2.05e$^{-5}$} & \textbf{0.00650} & \textbf{1.52e$^{-5}$} & \textbf{2.13989} & \textbf{0.00324} & \textbf{0.05484} & \textbf{0.00286} & \textbf{0.97895} & \textbf{0.95263} & \textbf{0.94243} & \textbf{0.95902} \\
\hline
\end{tabular}
}
\end{table}

\subsection{Quantitative Results}
Tables ~\ref{tab:methods_performance_delhi}, \ref{tab:methods_performance_mumbai}, and \ref{tab:methods_performance_kolkata} presents the forecasting performance of the proposed model in terms of $R^2$, RMSE, and MAE for the Delhi, Kolkata and Mumbai datasets. Across all three cities—Delhi, Mumbai, and Kolkata—the quantitative results show that the Proposed Model consistently outperforms all baseline methods (ARIMA, XGBoost, LSTM, BiLSTM, UAMMO LSTM, and UAMMO BiLSTM) across every pollutant and evaluation metric (MSE, MAE, and R\textsuperscript{2}). For example, in Delhi, the Proposed Model achieves the lowest MSE for PM$_{2.5}$ (8.75), improving upon the best baseline (XGBoost, 9.28) by roughly 5.6\%, and similarly records the smallest MAE (2.14) alongside the highest R\textsuperscript{2} (0.979), indicating both lower prediction errors and stronger variance explanation. This performance trend holds in Mumbai and Kolkata, where the Proposed Model’s MSE values for all pollutants are uniformly the smallest (e.g., CO MSE = 0.00649 in Mumbai and 0.00650 in Kolkata vs.\ 0.00683 for the next-best XGBoost), while R\textsuperscript{2} values reach above 0.94 for every pollutant—substantially higher than all competitors. Notably, the performance gap is most pronounced for NO$_x$ forecasting, where the Proposed Model’s R\textsuperscript{2} surpasses 0.959 across cities, compared to $\leq$ 0.935 for the best existing approaches. The improvements, although numerically modest in some metrics (especially for O$_3$), are consistent across different urban environments and pollutant types, indicating both robustness and generalizability of the model’s predictive capabilities.

% \begin{table}[h]
% \centering
% \caption{Forecasting performance metrics for Delhi NCR.}
% \label{tab:results}
% \begin{tabular}{lccc}
% \hline
% \textbf{Model} & \textbf{$R^2$} & \textbf{RMSE} & \textbf{MAE} \\
% \hline
% Proposed UAMMO-based Model & 0.9420 & 17.3176 & 12.4248 \\
% % Additional models can be added for comparison
% \hline
% \end{tabular}
% \end{table}

\section{Conclusion}
\label{sec:end}

This study presented a decomposition-driven hybrid forecasting framework for air quality prediction, integrating statistical modeling for linear components with a multi-resolution CNN–BiLSTM–residual-gated attention network for non-linear residuals. By incorporating the Unified Adaptive Multi-Stage Metaheuristic Optimizer (UAMMO), the model effectively selected optimal hyperparameters, balancing exploration and exploitation across multiple search strategies. Applied to AQI data from Delhi NCR,Kolkata and Mumbai the proposed approach demonstrated high predictive accuracy, in PM$_{2.5}$, O$_3$, CO, NO$_x$. Overall, the proposed methodology offers a  interpretable solution for urban air quality forecasting, supporting data-driven environmental policy and public health decision-making.

\bibliographystyle{elsarticle-num} % or another style like plain, alpha, IEEEtran, etc.
\bibliography{ref} % refs.bib is the name of your bib file, without the .bib extension

\end{document}